\documentclass[10pt,twocolumn,letterpaper]{article}

\usepackage{wacv}
\usepackage{times}
\usepackage{epsfig}
\usepackage{cite}
\usepackage{amsmath,graphicx}
\usepackage{epstopdf}
\usepackage{xcolor}
\usepackage{amssymb}
\usepackage{array}
\usepackage{booktabs}
\usepackage{multirow}
\usepackage{graphicx}
\usepackage{color}
\usepackage{float}
\usepackage{stfloats}
\usepackage[misc]{ifsym}
\usepackage{stmaryrd}
\usepackage{grffile}
\usepackage{subfig}
\usepackage{makecell}
\usepackage{float}
\usepackage{threeparttable}
\usepackage{enumitem}

\wacvfinalcopy 

\ifwacvfinal\fi

\begin{document}
\title{DeOccNet: Learning to See Through Foreground Occlusions in Light Fields}

\author{Yingqian~Wang$^{1}$, ~~~Tianhao~Wu$^{1}$, ~~~Jungang~Yang$^{1}$\thanks{\textit{Corresponding author}},\\
 Longguang~Wang$^{1}$, ~~~Wei~An$^{1}$, ~~~Yulan~Guo$^{1,2}$\\
 \\
$^{1}$College of Electronic Science and Technology, National University of Defense Technology, China\\
$^{2}$School of Electronics and Communication Engineering, Sun Yat-sen University, China\\
\tt\small \{wangyingqian16, yangjungang, yulan.guo\}@nudt.edu.cn}

\maketitle

\ifwacvfinal\thispagestyle{empty}\fi

\begin{abstract}
Background objects occluded in some views of a light field (LF) camera can be seen by other views. Consequently, occluded surfaces are possible to be reconstructed from LF images.  In this paper, we handle the LF de-occlusion (LF-DeOcc) problem using a deep encoder-decoder network (namely, \textit{DeOccNet}). In our method, sub-aperture images (SAIs) are first given to the encoder to incorporate both spatial and angular information. The encoded representations are then used by the decoder to render an occlusion-free center-view SAI. To the best of our knowledge, \textit{DeOccNet} is the first deep learning-based LF-DeOcc method. To handle the insufficiency of training data, we propose an LF synthesis approach to embed selected occlusion masks into existing LF images. Besides, several synthetic and real-world LFs are developed for performance evaluation. Experimental results show that, after training on the generated data, our \textit{DeOccNet} can effectively remove foreground occlusions and achieves superior performance as compared to other state-of-the-art methods. Source codes are available at: \textcolor{red}{https://github.com/YingqianWang/DeOccNet}.
\end{abstract}

\section{Introduction}
\label{sec:introduction}
Seeing through foreground occlusions is beneficial to many computer vision applications such as detection and tracking in surveillance \cite{joshi2007synthetic,pei2012novel,yang2011continuously,yang2013new}. However, due to foreground occlusions, some rays cannot hit the sensors of traditional single-view cameras (\eg, Digital Single Lens Reflex). Therefore, objects behind occlusions cannot be fully observed and reliably reconstructed. In recent years, camera arrays \cite{wilburn2005high,wilburn2004high,vaish2004using,venkataraman2013picam,lin2015camera} have undergone a rapid development since they can record light fields (LFs) and provide a large number of viewpoints with rich angular information. The complementary information among different viewpoints is beneficial for the reconstruction of occluded surfaces since background objects occluded in some views can be seen by other views.

\begin{figure}[t]
\centering
\footnotesize
\begin{tabular}{cc}
\includegraphics[width=3.8cm]{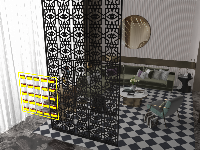} &
\includegraphics[width=3.8cm]{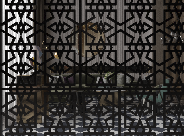}\\
(a)&
(b)\\
\includegraphics[width=3.8cm]{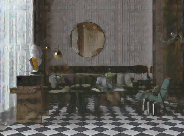} &
\includegraphics[width=3.8cm]{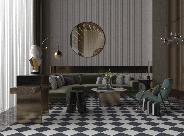} \\
(c) &
(d)\\
\end{tabular}
\caption{An illustration of LF-DeOcc using our rendered scenes \textit{Syn01}. (a) Configuration of the scene. Yellow boxes with $5\times5$ blocks represent camera arrays. (b) Occluded center-view SAI. (c) Results of our \textit{DeOccNet}. (d) Occlusion-free groundtruth.} \label{Thumbnail}
\end{figure}

As illustrated in Fig.~\ref{Thumbnail}, light field de-occlusion (LF-DeOcc) aims at removing foreground occlusions using sub-aperture images (SAIs) captured by a camera array\footnote{In the area of LF-DeOcc, images captured by camera arrays are widely used due to their wide baselines. Therefore, we follow the existing work \cite{vaish2004using,pei2013synthetic,yang2014all,xiao2017seeing,pei2018all} and use camera arrays for LF-DeOcc.}. The pioneering work on LF-DeOcc is proposed by Vaish \etal \cite{vaish2004using} using a refocusing method. However, this method cannot recover a clean surface of occluded objects since rays from occlusions and background are mixed. In fact, it is important but challenging to correctly select pixels only belonging to  occluded objects. To this end, existing methods \cite{pei2013synthetic,yang2014all,xiao2017seeing,pei2018all} generally built different models to handle LF-DeOcc problem. Due to the highly complex structures of scenes in real world, these methods with handcrafted feature extraction and stereo matching techniques cannot achieve satisfactory performance. In recent years, deep learning has been successfully used in different LF tasks such as depth estimation \cite{shin2018epinet, peng2018unsupervised}, image super-resolution \cite{zhang2019residual,wang2018lfnet,yeung2019light}, view synthesis \cite{wu2018light,wang2018end,wu2019learning}, and LF intrinsics \cite{alperovich2018light,alperovich2018intrinsic}. These networks have achieved state-of-the-art performance in numerous areas. However, to the best of our knowledge, deep learning has not been used for LF-DeOcc due to several issues. In this paper, we design a novel and effective paradigm, and propose the first deep learning network (\ie, \textit{DeOccNet}) to handle LF-DeOcc problem. Specifically, we summarize three major challenges in deep learning-based LF-DeOcc, and provide solutions to these challenges using our proposed paradigm.

The first challenge is that, as compared to LF depth estimation networks \cite{shin2018epinet, peng2018unsupervised} and LF super-resolution networks \cite{zhang2019residual,wang2018lfnet,yeung2019light}, LF-DeOcc networks should use as much information from occluded surfaces as possible, while maintaining a larger receptive field to cover occlusions of different types and scales. We address this challenge by employing an encoder-decoder network to encode LF structures. We concatenate all SAIs along the channel dimension to fully use the information of occluded surfaces. Besides, we use a residual atrous spatial pyramid pooling (ASPP) module to extract multi-scale features and enlarge receptive fields.

The second challenge is that, as compared to single image inpainting networks \cite{pathak2016context,yang2017high,yu2018generative,liu2018image}, LF-DeOcc networks have to learn the scene structure to automatically recognize, label and remove foreground occlusions. We address this challenge by setting the occlusion-free center-view SAI as groundtruth, and train our \textit{DeOccNet} in an end-to-end manner. In this way, our network can recognize occlusions from background through disparity discrepancy, and automatically remove foreground occlusions.

The third challenge is that, LF-DeOcc networks face an insufficiency of training data since large-scale LF datasets with removable foreground occlusions are unavailable. Moreover, test scenes are also insufficient for performance evaluation. We address this challenge by proposing a data synthesis approach to embed different occlusion masks into existing LF images. Using this approach, more than 1000 LFs are generated to train our network. Moreover, we develop several synthetic and real-world LFs for performance evaluation.

Experimental results have demonstrated the effectiveness of our paradigm. Our \textit{DeOccNet} achieves superior performance on both synthetic and real-world scenes as compared to other state-of-the-art methods.

\section{Related Works}
\label{sec:related work}

\subsection{Single image inpainting}
Single image inpainting methods aim at filling holes in an image using both neighborhood information and global priors. The major challenge of single image inpainting lies in synthesizing visually realistic and semantically plausible pixels for missing regions. Recent deep learning based methods \cite{pathak2016context,yang2017high,yu2018generative,liu2018image} have achieved promising results for inpainting large missing regions in an image. Specifically, Yu \etal \cite{yu2018generative} proposed a deep generative model-based inpainting method to synthesize novel image structures and textures. Liu \etal \cite{liu2018image} used partial convolutions for inpainting with irregular holes and achieved the state-of-the-art performance.

Compared to single image inpainting, LF-DeOcc can use complementary information provided by SAIs to generate improved results. The difference between single image inpainting and LF-DeOcc is significant. In single image inpainting, holes or masks are always pre-defined. However, LF-DeOcc requires automatical extraction of foreground occlusions by analyzing scene structures. That is, occlusions are closer to cameras than background objects, and thus have larger disparities. Due to the complex structures of real-world scenes, it is highly challenging for algorithms to correctly select rays originating from occluded objects.

\begin{figure*}[t]
\centering
\includegraphics[width=13cm]{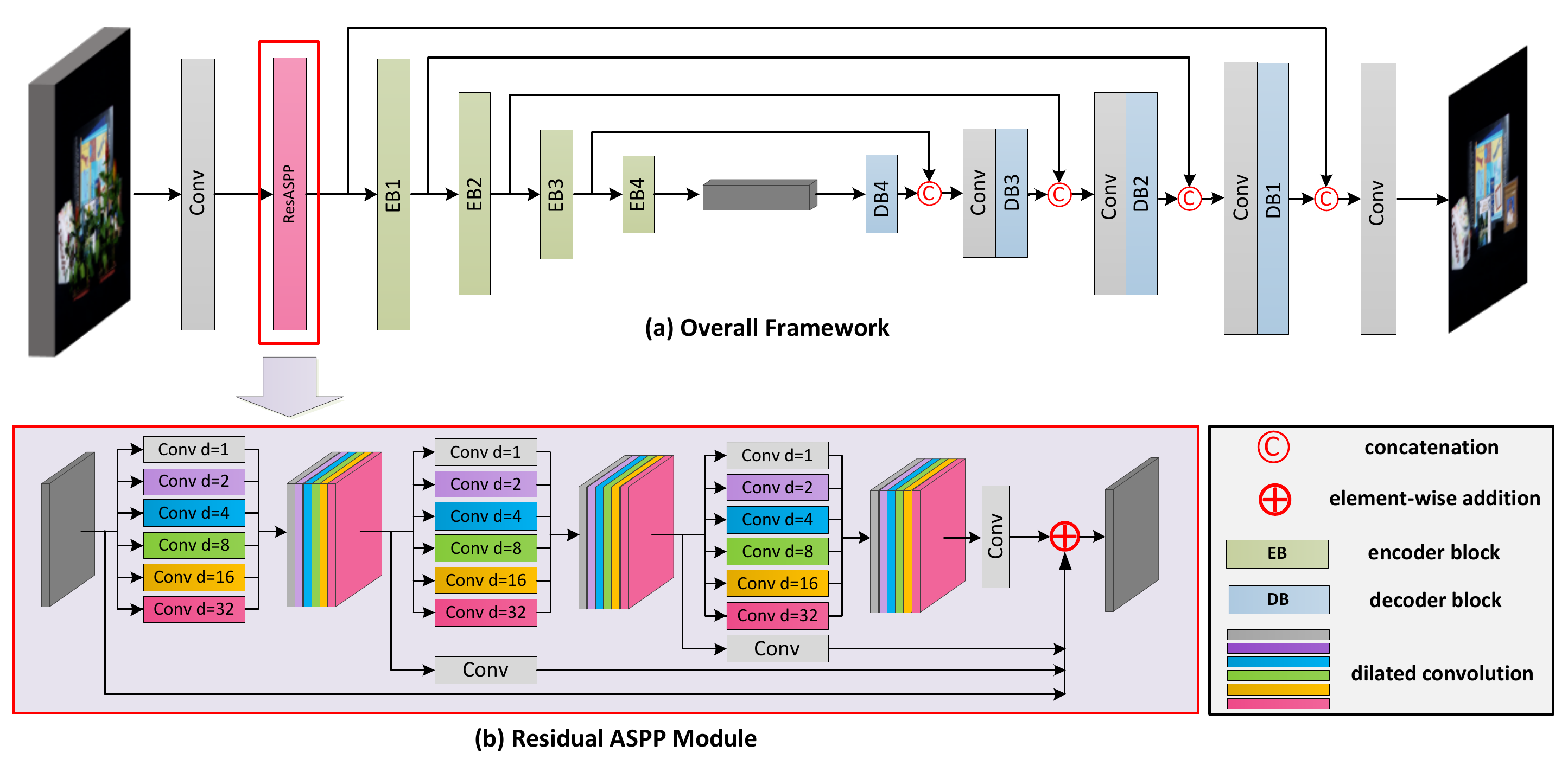}
\caption{{An overview of our \textit{DeOccNet}. (a) The overall architecture. (b) The structure of the residual ASPP module.}
\label{DeOccNet}}
\end{figure*}

\subsection{Light field de-occlusion}
LF-DeOcc is an active research topic and has been investigated for decades \cite{vaish2004using,vaish2006reconstructing,pei2013synthetic,yang2014all,xiao2017seeing,pei2018all}. Vaish \etal \cite{vaish2004using} proposed a refocusing method by warping each SAI by a specific value, and then averaging the warped SAIs along angular dimension. Due to the large equivalent aperture of camera arrays, when background is refocused on, occlusions are extremely blurred and the \textit{see through} effect can be achieved. However, the resulting images are always blurred due to the indiscriminate use of rays from both occlusions and background. Vaish \etal further proposed an improved version using both median cost and entropy cost \cite{vaish2006reconstructing}. Since these methods \cite{vaish2004using,vaish2006reconstructing} do not exactly exploit scene structures, their performance is limited for scenes with heavy occlusions.

To solve this problem, Pei \etal proposed a pixel-labeling method \cite{pei2013synthetic} to remove occlusions. Specifically, pixels corresponding to occlusions are labeled by stereo matching and masked out during refocusing process, resulting in a clean image. However, this method can only generate images refocused on a specific depth, leaving objects in other depth ranges suffering from various degrees of blurs. Subsequently, they used an image-matting approach to perform all-in-focus synthetic aperture imaging \cite{pei2018all}. Besides, Yang \etal used visible layers \cite{yang2014all} to address the all-in-focus imaging issue. Xiao \etal \cite{xiao2017seeing} used k-means clustering to classify pixels of occlusions and background. All these methods \cite{pei2013synthetic,yang2014all,xiao2017seeing,pei2018all} use handcrafted feature extraction and stereo matching techniques, and cannot achieve a satisfactory performance in scenes with complex structures and heavy occlusions.

\subsection{Deep learning in light field}

Deep neural networks have been widely used in various LF tasks such as image super-resolution \cite{yoon2015learning,wang2018lfnet,yeung2019light,zhang2019residual}, view synthesis \cite{wu2018light,wang2018end,wu2019learning}, and depth estimation \cite{shin2018epinet,peng2018unsupervised}. Compared to these tasks, networks for LF-DeOcc should have a larger receptive field and use more information of occluded surfaces. Currently, no existing work on deep learning based LF-DeOcc is available in literature. It is worth noting that works in \cite{alperovich2018light,alperovich2018intrinsic} are similar to ours. Specifically, a fully convolutional auto-encoder is proposed in \cite{alperovich2018light} to separate diffuse and specular components of an LF. Both \cite{alperovich2018light} and our work require high-level features and global priors of scene structures. Consequently, we built our network upon encoder-decoder architecture to encode LF structures. Note that, there are two significant differences between \cite{alperovich2018light} and our network. First, only horizontal and vertical SAIs (\eg, 9 cross-views in a 5$\times$5 LF) are used in \cite{alperovich2018light}. In contrast, all SAIs are used in our network to fully exploit the information of occluded objects. Second, our \textit{DeOccNet} uses residual ASPP module to enlarge the receptive field, and uses multiple skip layers to have a holistic understanding of the scene while preserving fine details.

\section{The Proposed Method}
\label{sec:method}
\subsection{Network architecture}

The task of our \textit{DeOccNet} is to replace pixels of occlusions with pixels from the background. To achieve this task, our network is required to find correspondence and incorporate complementary information from SAIs. Note that, foreground occlusions generally have shallow depths and large disparities. That is, pixels of occlusions always have very large position variations among SAIs. Therefore, multi-scale features with large receptive fields are required for our network. In this paper, we first use a residual ASPP module for hierarchical feature extraction, and then use an auto-encoder to incorporate both spatial and angular information. The architecture of our \textit{DeOccNet} is shown in Fig.~\ref{DeOccNet}. Different from existing LF networks \cite{alperovich2018light,shin2018epinet,zhang2019residual} where only part of SAIs are stacked as inputs, we stacked all SAIs along the channel dimension (\eg, $75$ channels for $5\times5$ RGB SAIs) to use as much information as possible since LF-DeOcc highly depends on the information of occluded objects. Consequently, our \textit{DeOccNet} takes the stacked SAIs as its input, and finally generates an occlusion-free center-view SAI.

\textbf{Residual ASPP module}. In our network, the input volume is first processed by a $1\times1$ convolution layer to generate features with a fixed depth (\ie, $64$ in this paper). Then, a residual ASPP module is used to generate hierarchical features. As shown in Fig.~\ref{DeOccNet}(b), similar to the module used in \cite{wang2019learning}, we first combine six dilated convolution layers (with dilation rates of $1$, $2$, $4$, $8$, $16$, and $32$) to form an ASPP group, then cascade three ASPP groups in residual manner to achieve high learning efficiency. The residual ASPP module can enlarge the receptive field and extract multi-scale information around occlusions. It is demonstrated in the ablation study (see Table \ref{tab1}) that our residual ASPP module is beneficial to the overall LF-DeOcc performance.

\begin{figure}
\centering
\includegraphics[width=8cm]{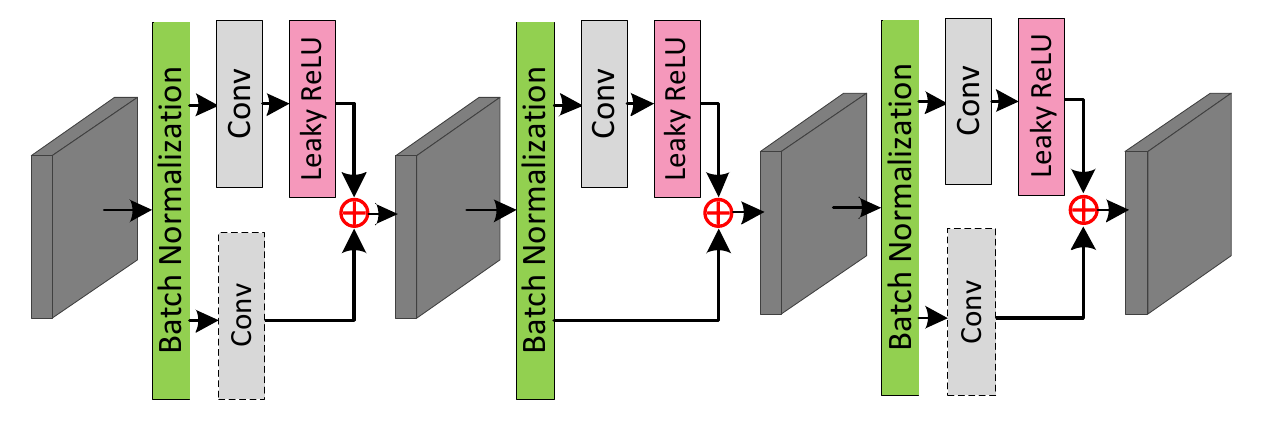}
\caption{{The structure of the encoder and decoder blocks. Note that, the encoder and decoder blocks share mirrored structures. That is, strided convolution is used in the third unit in each encoder block, while de-convolution is used in the first unit in each decoder block.}
\label{EBDB}}
\vspace{-0.3cm}
\end{figure}

\begin{figure*}[t]
\centering
\begin{tabular}{cc}
\includegraphics[height=3.5cm]{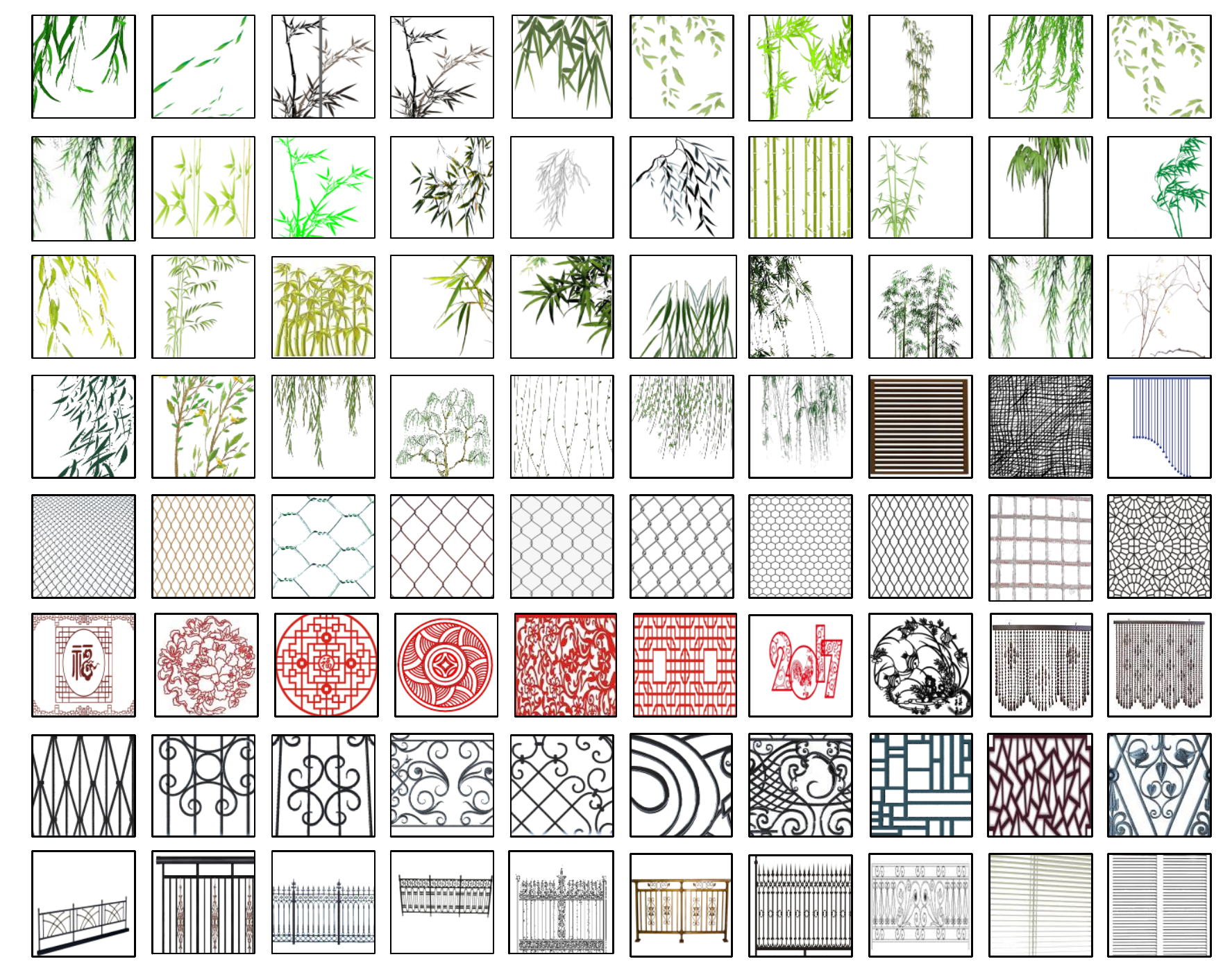} &
\includegraphics[height=3.6cm]{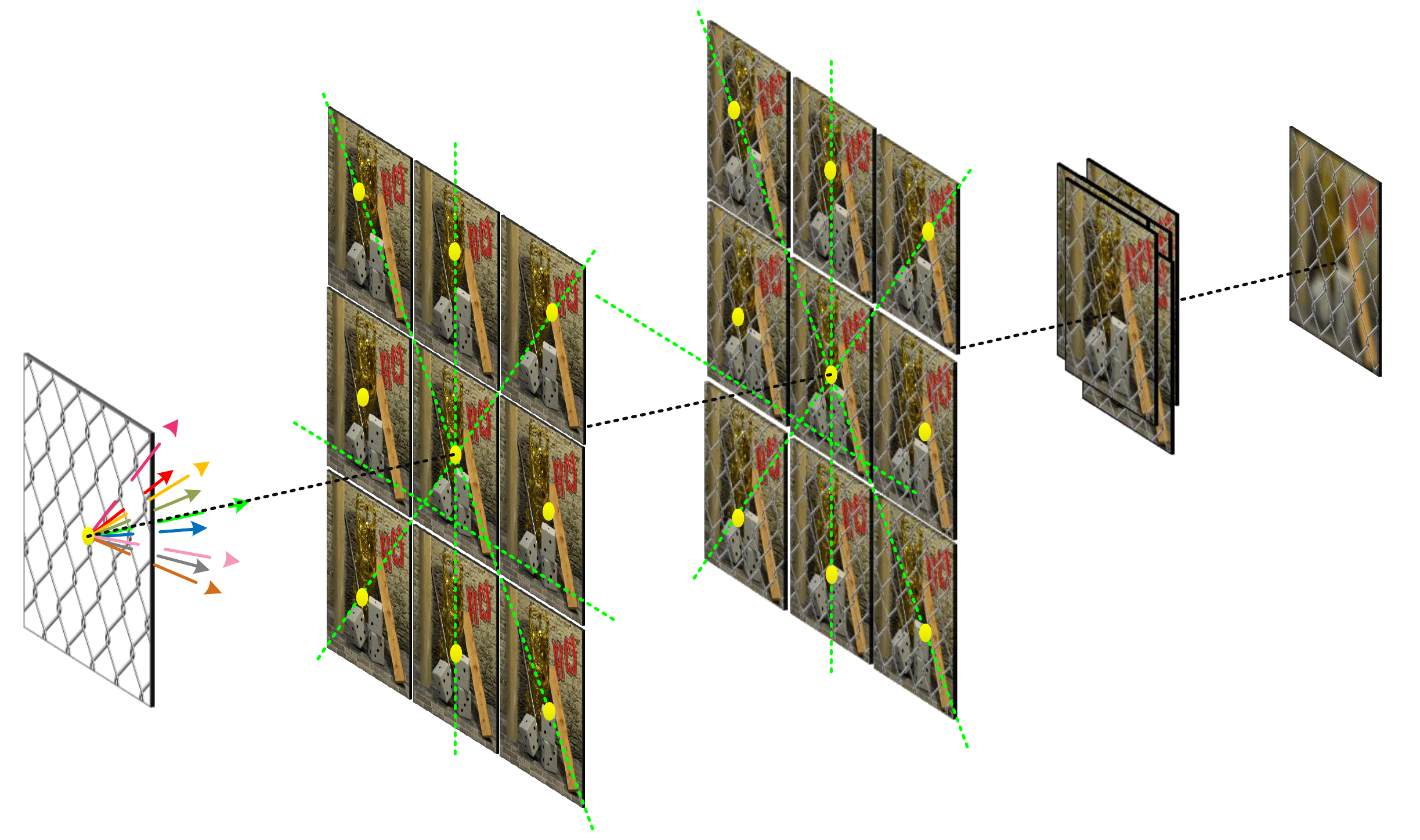}\\
(a)&
(b)\\
\end{tabular}
\vspace{-0.1cm}
\caption{{An illustration of our \textit{Mask Embedding} approach. (a) Masks used in our approach. Note that, cropping and scaling are performed for better visualization. (b) The pipeline of our \textit{Mask Embedding} approach. Here, a $3\times3$ LF is used as an example.}
\label{MaskEmbedding}}
\end{figure*}
\textbf{Encoder pathway}. Features generated by the residual ASPP module are then transferred to the encoder pathway, where $4$ encoder blocks are cascaded to incorporate both spatial and angular information. Specifically, as shown in Fig.~\ref{EBDB}, each encoder block contains three cascaded units. In each unit, the batch-normalized features are given to two separated paths to achieve local residual learning. The first path includes a $3\times3$ convolution and a Leaky ReLU (with a leaky factor of $0.1$), while the second path either keeps the input unchanged or passes it through a strided convolution or de-convolution. Therefore, features produced by the two paths have the same resolution. Features from both paths are added to produce the output of this unit. For each encoder block, the first two units keep depth and resolution unchanged, while the third unit applies a strided convolution (with a stride of $2$) to halve the resolution and double the feature depth. Consequently, the final feature generated by the encoder (bottleneck of our \textit{DeOccNet}) is $1/16$ time of the input feature in resolution and has $1024$ channels in depths.

\textbf{Decoder pathway}. After passing through the bottleneck, features are decoded through the decoder pathway. Note that, decoder blocks have mirrored structures as encoder blocks. That is, the decoder block is also chained by three residual units, and the first unit uses a de-convolution to exactly revert the encoder block on the corresponding level. Moreover, to preserve fine details in the final output image, features of different resolutions on the encoder pathway are concatenated with their counterparts on the decoder pathway by skip connections. Consequently, the decoder can be guided to gradually add details onto restored images. Since feature depth is doubled by concatenation, an additional $1\times1$ convolution layer is employed before the last three decoder blocks to halve the feature depth. Similarly, a $1\times1$ convolution layer is also applied to the output feature to reduce its channel to $3$.

\subsection{Mask embedding for training data synthesis}

It is important to provide sufficient data to train our \textit{DeOccNet}. Although LFs with removable occlusions can be acquired by capturing real-world scenes with/without foreground occlusions, or by rendering synthetic scenes using softwares such as \textit{3dsMax}\footnote{https://www.autodesk.eu/products/3ds-max/overview} and \textit{Blender}\footnote{https://www.blender.org/}, these approaches are significantly labour-intensive and even infeasible. Consequently, it is important to design an efficient approach to generate a large amount of data for network training. In this paper, we propose \textit{Mask Embedding}, a training data synthesis approach to synthesize LFs with removable foreground occlusions. An illustration of our \textit{Mask Embedding} approach is shown in Fig.~\ref{MaskEmbedding}.

As shown in Fig.~\ref{MaskEmbedding}(a), we manually collected $80$ mask images from Internet using tags such as salix leaves, grids, fences, and paper cuts. All these masks are common foreground occlusions in daily life. Meanwhile, we collected $60$ LFs from the Stanford LF dataset \cite{vaish2008new}, the Old HCI 4D LF dataset \cite{wanner2013datasets}, the New HCI 4D LF benchmark \cite{honauer2016dataset}, and the MIT Synthetic LF Archive \cite{lanman2011polarization}. Note that, to improve the generalization capability of our network, the RGB channels of both LFs and mask images are randomly shuffled, and the masks are randomly selected to be embedded into shuffled LFs.

The pipeline of our \textit{Mask Embedding} approach is illustrated in Fig.~\ref{MaskEmbedding}(b). We randomly pick a mask from the mask set and then embed it into each SAI according to LF configurations. Specifically, a disparity with a shallow depth is randomly set and allocated to the selected mask. Then, the mask is warped according to the angular coordinate of the target view and the allocated disparity. Note that, bilinear interpolation is used when masks do not fall into integer coordinates. Next, warped masks are added to each SAI, resulting in LFs with foreground occlusions. Finally, refocusing is performed on each generated LF to check the LF configuration.

\begin{figure*}[t]
\centering
\footnotesize
\begin{tabular}{cccc}
\includegraphics[width=3.7cm]{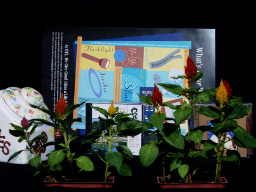} &
\includegraphics[width=3.7cm]{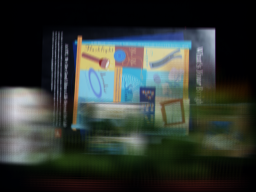} &
\includegraphics[width=3.7cm]{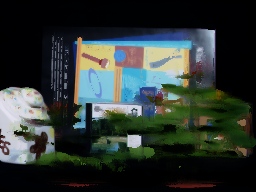} &
\includegraphics[width=3.7cm]{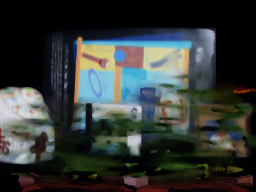} \\
(a) Occluded &
(b) Refocus \cite{vaish2004using} &
(c) Median \cite{vaish2006reconstructing} &
(d) Pei \etal \cite{pei2018all}\\
\includegraphics[width=3.7cm]{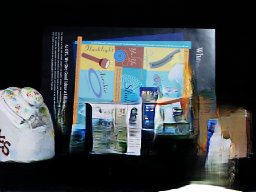} &
\includegraphics[width=3.7cm]{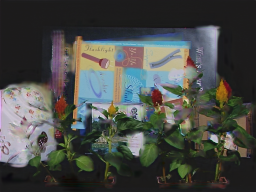} &
\includegraphics[width=3.7cm]{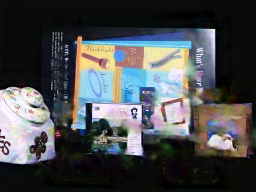} &
\includegraphics[width=3.7cm]{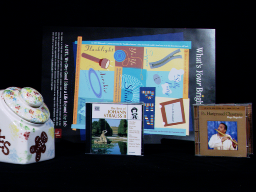} \\
(e) Liu \etal \cite{liu2018image} &
(f) \textit{DeOccNet} (75 center views)&
(g) \textit{DeOccNet} &
(h) Groundtruth \\
\end{tabular}
\caption{Qualitative results achieved on the CD scene \cite{vaish2008new} (with an occlusion rate of $40.2\%$). (a) Occluded center-view SAI. (b)-(e) Comparative results achieved by different methods. (f) Result achieved by our \textit{DeOccNet} using $75$ identical center-view SAIs as its inputs (discussed in Section~\ref{discussion}). (g) Our result. (h) Occlusion-free center-view SAI.}\label{CDScenes}
\vspace{-0.1cm}
\end{figure*}

Although the \textit{Mask Embedding} approach can easily generate a large number of LFs for training, the generated LFs only have occlusions in a single depth, which is significantly different from real-word scenarios. To address this issue, we use generated LFs as original LFs and repeat this process twice to synthesize LFs with occlusions at two and three depth layers. Using the proposed approach, we totally synthesize $1500$ LFs to train our models. Although LFs synthesized by our approach only have fence-like and front-parallel occlusions, experimental results show that, our network trained on the synthetic data can generalize well to real-world cases (\eg, CD scene in Fig.~\ref{CDScenes}). That is, our network can successfully learn the scene structure through disparity discrepancy using LFs synthesized by our \textit{Mask Embedding} approach.

\subsection{Training details}

We trained two models with $5\times15$ and $5\times5$ input SAIs. The $5\times15$ model was used for the Stanford CD scene \cite{vaish2008new}, and the $5\times5$ model was used for our self-developed scenes. During the training phase, we used LFs synthesized by the \textit{Mask Embedding} approach as the training data. All the $60$ LFs were used for the $5\times5$ model. In contrast, due to the angular resolution limitation of existing LF datasets, only LFs from the Stanford LF dataset \cite{vaish2008new} were used to generate training data for the $5\times15$ model.

\begin{figure}[t]
\centering
\includegraphics[width=8.3cm]{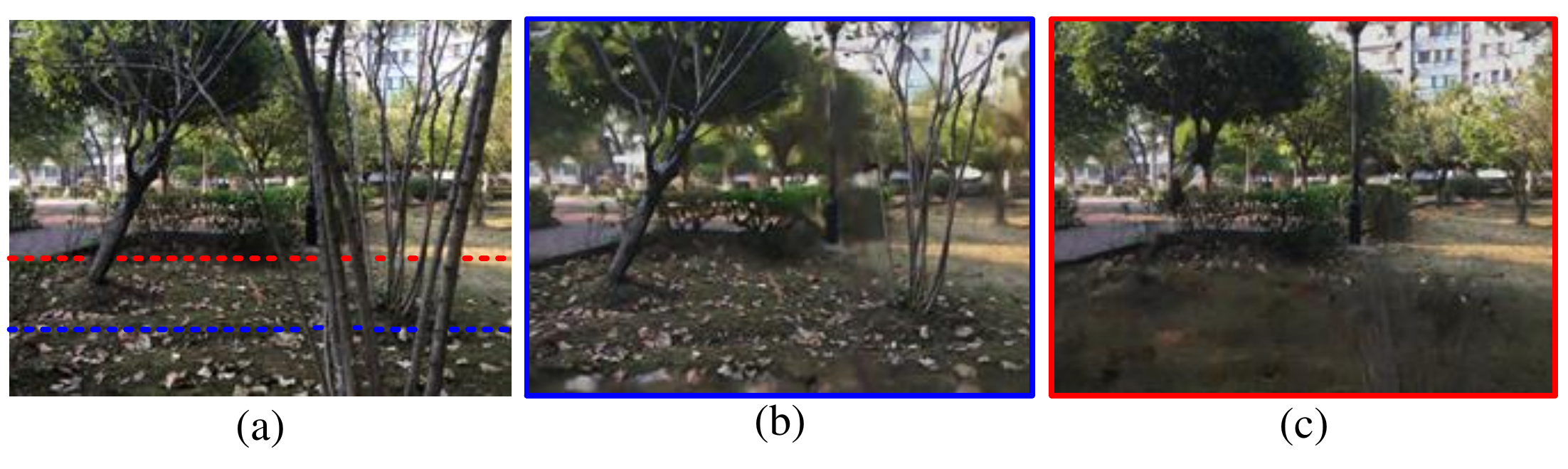}
\caption{Multi-occlusion situation in LF-DeOcc. (a) Occluded center-view SAI. (b) Result of our \textit{DeOccNet} with inputs being rectified at a shallow depth (the blue dotted line in (a)). The front-most tree is considered as a foreground occlusion. (c) Result of our \textit{DeOccNet} with inputs being rectified at a deep depth (the red dotted line in (a)). Three front trees are considered as foreground occlusions. Consequently, different results can be generated by our \textit{DeOccNet} with the same inputs being rectified at different depth values.}
\label{TwoDepth}
\vspace{-0.2cm}
\end{figure}

It is worth noting that foreground and background in a scene are relative concepts. That is, some occlusions can also be considered as background objects in multi-occlusion situations as shown in Fig.~\ref{TwoDepth}. Consequently, both training and test scenes should be rectified to specific depths for LF-DeOcc. In this paper, we perform rectification by cropping each SAI accordingly to make occlusions have positive disparity values while backgrounds have negative disparity values. In this way, our \textit{DeOccNet} can effectively achieve LF-DeOcc by simply removing objects with positive disparity values. Finally, we cropped occluded SAIs into $224\times224$ pixel patches by a stride of $112$, and performed $2\times$ upsampling for data augmentation. Meanwhile, occlusion-free center-view SAI was cropped and upsampled accordingly to generate groundtruths.

Our \textit{DeOccNet} was implemented in Pytorch on a PC with an Nvidia RTX 2080Ti GPU. All models were trained using an MSE loss and optimized using the Adam method \cite{kingma2014adam} with $\beta_{1}=0.9$, $\beta_{2}=0.999$, and a batch size of $8$. The initial learning rate was set to $1\times10^{-3}$ and reduced to $1\times10^{-4}$ after $100$ epochs. The training took about $2$ days and was stopped at $200$ epochs.

\section{Experiments}
\label{sec:experiment}
In this section, we first introduce the test scenes used in our experiments, and then compare our method to several state-of-the-art methods. Finally, we present ablation study and analyses.

\subsection{Test scenes}

\textbf{Real-world scenes.} We followed \cite{pei2018all,vaish2006reconstructing,yang2014all} and tested our method on the publicly available CD scene \cite{vaish2008new}. The original CD scene consists of 105 views distributed on a $5\times21$ grid. We selected the central $5\times15$ views for performance evaluation. Groundtruth image is provided by a second capture with occlusions being removed. Besides, we captured several real-world scenes using a moving \textit{Leica Q} camera (with an $F=10$, $f=28$ \textit{mm} lens) mounted on a gantry. It is argued in \cite{wu2017light,wang2018selective} that the scanning scheme is equivalent to a single shot by a camera array in static occasions. We shifted the camera to $25$ positions on a $5\times5$ grid with $3$ \textit{cm} baselines. The captured images were calibrated using the method in \cite{zhang2000flexible}.

\textbf{Synthetic scenes.} Since the number of real-world test scenes is very small, we rendered $4$ synthetic scenes with removable foreground occlusions for further evaluation. All elements in our synthetic scenes were collected from Internet, and parameters (\eg, lighting, depth range) were tuned to better reflect real scenes. The angular resolution of each scene was set to $5\times5$, while baselines and occlusion ranges were varied in different scenes. Occlusion-free center-view SAIs were also rendered for quantitative evaluation.

\begin{figure*}[t]
\centering
\footnotesize
\begin{tabular}{ccccc}
\includegraphics[width=3cm]{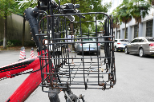} &
\includegraphics[width=3cm]{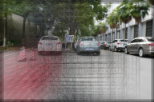} &
\includegraphics[width=3cm]{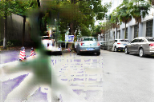} &
\includegraphics[width=3cm]{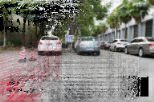} &
\includegraphics[width=3cm]{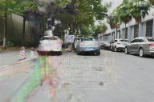}\\
(a) Occluded (Bike01)&
(b) Refocus \cite{vaish2004using}&
(c) Liu \etal \cite{liu2018image} &
(d) Pei \etal \cite{pei2018all} &
(e) Ours \\[1pt]
\includegraphics[width=3cm]{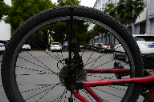} &
\includegraphics[width=3cm]{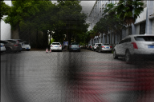} &
\includegraphics[width=3cm]{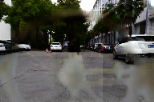} &
\includegraphics[width=3cm]{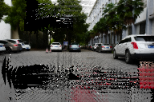} &
\includegraphics[width=3cm]{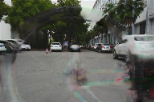}\\
(f) Occluded (Bike02)&
(g) Refocus \cite{vaish2004using}&
(h) Liu \etal \cite{liu2018image} &
(i) Pei \etal \cite{pei2018all} &
(j) Ours \\[1pt]
\includegraphics[width=3cm]{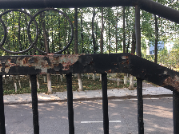} &
\includegraphics[width=3cm]{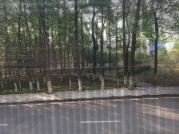} &
\includegraphics[width=3cm]{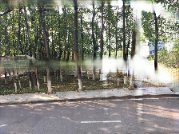} &
\includegraphics[width=3cm]{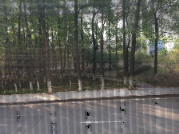} &
\includegraphics[width=3cm]{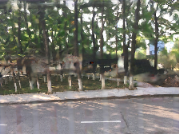}\\
(k) Occluded (Handrail)&
(l) Refocus \cite{vaish2004using}&
(m) Liu \etal \cite{liu2018image} &
(n) Pei \etal \cite{pei2018all} &
(o) Ours \\[1pt]
\end{tabular}
\caption{Qualitative results achieved on our self-developed real-world scenes. Note that, the occlusion rates of the three scenes are $61.8\%$, $57.7\%$, and $39.1\%$, respectively.}\label{BikeScenes}
\end{figure*}

\begin{figure*}[t]
\centering
\footnotesize
\begin{tabular}{ccccc}
\includegraphics[width=3cm]{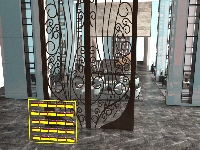} &
\includegraphics[width=3cm]{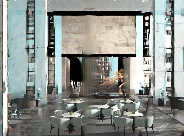} &
\includegraphics[width=3cm]{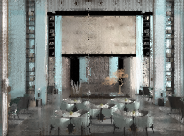} &
\includegraphics[width=3cm]{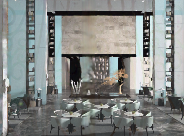} &
\includegraphics[width=3cm]{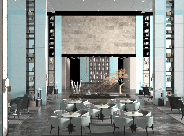}\\
(a) Overview (Syn02)&
(b) Liu \etal \cite{liu2018image} &
(c) Pei \etal \cite{pei2018all} &
(d) Ours &
(e) Groundtruth\\
\includegraphics[width=3cm]{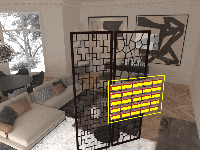} &
\includegraphics[width=3cm]{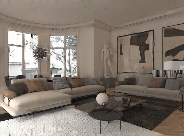} &
\includegraphics[width=3cm]{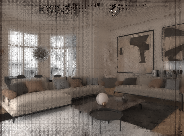} &
\includegraphics[width=3cm]{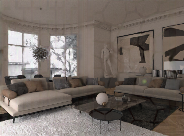} &
\includegraphics[width=3cm]{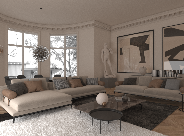}\\
(f) Overview (Syn03)&
(g) Liu \etal \cite{liu2018image} &
(h) Pei \etal \cite{pei2018all} &
(i) Ours &
(j) Groundtruth\\
\includegraphics[width=3cm]{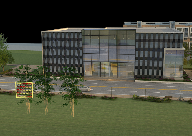} &
\includegraphics[width=3cm]{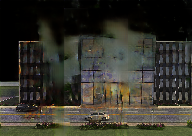} &
\includegraphics[width=3cm]{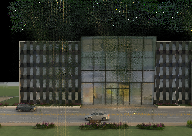} &
\includegraphics[width=3cm]{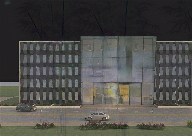} &
\includegraphics[width=3cm]{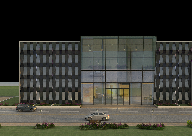}\\
(k) Overview (Syn04)&
(l) Liu \etal \cite{liu2018image} &
(m) Pei \etal \cite{pei2018all} &
(n) Ours &
(o) Groundtruth\\
\end{tabular}
\caption{Qualitative results achieved on our synthetic scenes. Sub-figures on the leftmost column represent the configurations of different scenes, yellow boxes with $5\times5$ blocks represent camera arrays. Note that, the occlusion rates of these scenes are $30.2\%$, $19.6\%$ and $14.8\%$, respectively.} \label{SynScenes}
\vspace{-0.2cm}
\end{figure*}
\subsection{Comparison to the state-of-the-arts}

We compared our \textit{DeOccNet} to the state-of-the-art LF-DeOcc method \cite{pei2018all}. We also used traditional refocusing method \cite{vaish2004using} and its improved version \cite{vaish2006reconstructing} as baselines. Moreover, to investigate the benefits of complementary information introduced by additional perspectives, we compared our method to the state-of-the-art image inpainting method \cite{liu2018image}. Note that, the inpainting method \cite{liu2018image} cannot automatically recognize occlusions in an image. Therefore, we manually labeled occlusions in each center-view image. Since the codes of \cite{pei2018all,vaish2004using,vaish2006reconstructing} are unavailable, we used our own implementations with their default parameter settings. Following the state-of-the-art image inpainting methods \cite{yu2018generative,liu2018image}, mean $l_1$ error, peak signal-to-noise ratio (PSNR) and structure similarity (SSIM) are used as quantitative evaluation metrics in this paper. Readers are referred to \cite{yu2018generative,liu2018image} for more details about these metrics. Qualitative results on real-world and synthetic datasets are shown in Figs. \ref{CDScenes}, \ref{BikeScenes}, and \ref{SynScenes}, and quantitative results are listed in Table \ref{tab1}.

It can be seen from Fig. \ref{CDScenes}(a) that occlusions in the CD scene occupy a large portion. Consequently, even though $75$ views are provided, it is still highly challenging to reconstruct occluded objects. Method \cite{vaish2004using} removes occlusions by warping and averaging SAIs. As shown in Fig. \ref{CDScenes}(b), although occlusions are extremely blurred at the focused depth, occluded object is also unclear and its contrast is low. Besides, since results produced by \cite{vaish2004using} only focus on limited depth ranges, objects in other depth ranges are highly blurred and cannot be recognized. In contrast, Vaish \etal \cite{vaish2006reconstructing} use median cost to achieve all-in-focus synthetic aperture imaging. However, the performance achieved by \cite{vaish2006reconstructing} is very limited.

Pei \etal \cite{pei2018all} achieve a better performance than method \cite{vaish2006reconstructing}. That is because, only rays belonging to occluded objects are used in \cite{pei2018all}. However, due to various textures and shapes of occlusions and background, it is difficult to exactly select rays only from background, and the resulting images can be deteriorated by incorrect classification of occlusion and background. More specifically, if pixels from occlusions are misused, the resulting images will be blurred (\eg, Fig. \ref{CDScenes}(d)). In contrast, if background are incorrectly considered as occlusions, the resulting images will have blank areas (\eg, Figs. \ref{BikeScenes}(d) and \ref{BikeScenes}(i)) since all information in these areas is removed.

The single image inpainting method \cite{liu2018image} tries to hallucinate missing part and render reasonable results using knowledge learned from a large number of daily scenes (\eg, the \textit{ImageNet} database \cite{deng2009imagenet}). It can generate promising results on scenes with simple structure and small occlusions (\eg, Fig. \ref{SynScenes}(g)). However, since no additional information is used in the inpainting process, this method cannot handle scenes with rich textures/details or heavy occlusions (\eg, Figs. \ref{BikeScenes}(c)and \ref{SynScenes}(l)).

As compared to existing methods, our method achieves superior performance, especially on scenes with heavy occlusions (\eg, scenes CD and Syn04). That is, \textit{DeOccNet} successfully learns to discriminate occlusions from backgrounds, and incorporates information of occluded objects from different viewpoints. It can be also seen from Table \ref{tab1} that our method achieves the best quantitative results on scenes CD, Syn01 and Syn02. Note that, our method is slightly inferior to \cite{liu2018image} on scene Syn03. That is because, method \cite{liu2018image} uses manually labeled groundtruth occlusions while our method do not rely on any groundtruth information. Since scene Syn03 has relatively small occlusions and simple textures, method \cite{liu2018image} works well using learned spatial priors and labeled groundtruth occlusions.

\begin{table*}[t]

\centering
\footnotesize
\caption{Quantitative results achieved by different methods and different design choices of \textit{DeOccNet}. Note that, for $l_1$-error, smaller scores indicate better performance, and for \textit{PSNR} and \textit{SSIM}, higher scores indicate better performance.}\label{tab1}
\vspace{-0cm}
\begin{tabular}{|>{}m{3cm}|>{\centering}m{1.5cm}>{\centering}m{1.5cm}>{\centering}m{1.5cm}
>{\centering}m{1.5cm}>{\centering}m{1.5cm}>{\centering}m{1.5cm}|}
\hline
~ & CD &  Syn01 & Syn02 & Syn03 & Syn04 & Average\tabularnewline
\hline
$l_1$-error (Pei \etal \cite{pei2018all}) & 0.233 & 0.188 & 0.240 & 0.204 & \textbf{0.156} & 0.204 \tabularnewline
$l_1$-error (Liu \etal \cite{liu2018image}) & 0.196 & 0.264 & 0.180 & \textbf{0.078} & 0.187 & 0.181 \tabularnewline
$l_1$-error (ours\_noASPP) & 0.296 & 0.200 & 0.271 & 0.220 & 0.339 & 0.265 \tabularnewline
$l_1$-error (ours\_3skips) & 0.222 & 0.202 & 0.346 & 0.319 & 0.207 & 0.259 \tabularnewline
$l_1$-error (ours) & \textbf{0.185} & \textbf{0.138} & \textbf{0.165} & 0.163 & 0.242 & \textbf{0.178} \tabularnewline
\hline
\textit{PSNR} (Pei \etal \cite{pei2018all}) & 16.75 & 19.78 & 18.10 & 19.60 & \textbf{21.57} & 19.16 \tabularnewline
\textit{PSNR}(Liu \etal \cite{liu2018image}) & 15.95 & 18.19 & 18.99 & \textbf{25.58} & 20.44 & 19.83 \tabularnewline
\textit{PSNR} (ours\_noASPP) & 18.80 & 20.37 & 17.69 & 20.78 & 17.79 & 19.09 \tabularnewline
\textit{PSNR} (ours\_3skips) & 19.95 & 20.41 & 16.25 & 17.74 & 21.20 & 19.11 \tabularnewline
\textit{PSNR} (ours) & \textbf{21.27} & \textbf{24.68} & \textbf{21.74} & 23.98 & 20.56 & \textbf{22.45} \tabularnewline
\hline
\textit{SSIM} (Pei \etal \cite{pei2018all}) & 0.508 & 0.636 & 0.568 & 0.656 & 0.569 & 0.587 \tabularnewline
\textit{SSIM} (Liu \etal \cite{liu2018image}) & 0.647 & 0.595 & 0.682 & 0.848 & 0.485 & 0.651 \tabularnewline
\textit{SSIM} (ours\_noASPP) & 0.625 & 0.586 & 0.617 & 0.809 & 0.523 & 0.632 \tabularnewline
\textit{SSIM} (ours\_3skips) & 0.621 & 0.613 & 0.560 & 0.742 & 0.530 & 0.613 \tabularnewline
\textit{SSIM} (ours) & \textbf{0.694} & \textbf{0.699} & \textbf{0.734} & \textbf{0.858} & \textbf{0.650} & \textbf{0.727} \tabularnewline
\hline
\end{tabular}
\vspace{-0.2cm}
\end{table*}
\subsection{Ablation study}
We conducted ablation study to investigate the improvement introduced by the residual ASPP module and skip connections. Quantitative results are presented in Table \ref{tab1}.

First, we removed the residual ASPP module and retrained \textit{DeOccNet} from scratch using the same training data. We can observe from Table~\ref{tab1} that the residual ASPP module significantly contributes to the overall LF-DeOcc performance. Specifically, it introduces more than $3$ dB improvement in average PSNR, and nearly $0.1$ improvement in average SSIM. That is because, the residual ASPP module can help the network to have a large receptive field to cover foreground occlusions of various shapes and scales.

Then, we investigated the benefits introduced by skip connections. Note that, during the training process, we discovered that the network cannot achieve a reasonable convergence without any skip connection. Therefore, we only removed the outermost skip connection in the ablation study, and infer the contribution of skip layers from experimental results. As shown in Table~\ref{tab1}, the outermost skip layer introduces $3.34$ dB improvement in average PSNR and $1.14$ improvement in average SSIM. That is because, the high-frequency details are tend to be lost by strided convolutions, and it is difficult to recover these details from low-resolution features. Therefore, skip connections are necessary to provide a short path for low-level and high-frequency information.

\subsection{Further discussion}\label{discussion}
It is worth noting that LFs generated by our \textit{Mask Embedding} approach were used as the only training data of our \textit{DeOccNet}, and there is no intersection between the training and test scenes. In fact, the styles of occlusion and background between the training and test scenes differ significantly. That is, all occlusions in our training data are fence-like and front-parallel masks. Although we ran the \textit{Mask Embedding} approach three times to simulate occlusions at multiple depths, we cannot generate occlusions within a continuous range of depths as in the CD and Bike scenes. However, as shown in Figs.~\ref{CDScenes}(g) and \ref{BikeScenes}(e), our network can generalize well to these cases to handle slant occlusions. That is because, both training and test scenes share the same LF structures. Specifically, occlusions have positive disparity values while backgrounds have negative disparity values.

To investigate the intrinsic mechanism of our network in dealing with foreground occlusions, we stacked $75$ identical center-view SAIs of the CD scene, and fed them into our \textit{DeOccNet}. It can be seen from Fig. \ref{CDScenes}(f) that our \textit{DeOccNet} network cannot work with replicated inputs with identical information. That is, the mechanism of our \textit{DeOccNet} is significantly different from that of the single image inpainting network \cite{liu2018image}. Specifically, rather than using spatial information within one perspective in \cite{liu2018image}, our network uses complementary information from different viewpoints. It introduces significant performance improvements in scenes with heavy but pierced occlusions (\eg, basket in scene Bike01) because useful information can be introduced by different perspectives. However, only focusing on complementary information from different views makes our network perform unsatisfactorily on scenes with solid block regions (\eg, the right bottom corner in scene Bike02) since background is occluded in all perspectives and spatial information is need for LF-DeOcc. In the future, we will incorporate perspective information with neighborhood prior for LF-DeOcc, which is likely to introduce further performance improvement.

\section{Conclusion}
\label{sec:conclusion}
In this paper, we propose \textit{DeOccNet}, the first deep learning based method for LF-DeOcc. We embed masks into existing LFs to generate a large training dataset. Experiments on both synthetic and real-world scenes show that, our \textit{DeOccNet} can automatically remove foreground occlusions through disparity discrepancy, and achieve superior performance as compared to existing methods.

\section{Acknowlegement}
\label{sec:acknowlegement}
This work was partially supported by the National Natural Science Foundation of China (Nos. 61401474, 61602499, 61972435), and Fundamental Research Funds for the Central Universities (No. 18lgzd06).

{\small
\bibliographystyle{ieee}
\bibliography{WACV}
}

\end{document}